%% file: acl_latex.tex
\pdfoutput=1
\documentclass[11pt]{article}

\usepackage[final]{acl}

\usepackage{times}
\usepackage{latexsym}
\usepackage[linesnumbered,ruled,vlined]{algorithm2e} % 加载 algorithm2e 宏包
\usepackage{array} % 加载 array 宏包
\SetAlgoNlRelativeSize{0} % 去除行号                    
\SetAlgoNlRelativeSize{-2}
\SetAlgoNlRelativeSize{0}
\SetAlgoNlRelativeSize{-2} % 增加水平对齐样式

\usepackage[T1]{fontenc}
\usepackage[utf8]{inputenc}

\usepackage{microtype}

\usepackage{amsmath}
\usepackage{multicol}
\usepackage{multirow}
\usepackage{inconsolata}

\usepackage{graphicx}

\usepackage[normalem]{ulem}
\useunder{\uline}{\ul}{}

\usepackage{subcaption}    % 用于创建子图

\newcommand{\tref}[1]{Table~\ref{tab:#1}}
\newcommand{\fref}[1]{Figure~\ref{fig:#1}}

\newcommand{\eref}[1]{Equation~\ref{eq:#1}}

\title{XQuant: Achieving Ultra-Low Bit KV Cache Quantization with Cross-Layer Compression}

\author{
 \textbf{Haoqi Yang\textsuperscript{2}},
 \textbf{Yao Yao\textsuperscript{3}},
 \textbf{Zuchao Li\textsuperscript{1}\footnotemark[1]},
 \textbf{Baoyuan Qi\textsuperscript{4}},
 \textbf{Guoming Liu\textsuperscript{4}},
 \textbf{Hai Zhao\textsuperscript{3}}
\\
 \textsuperscript{1}School of Artificial Intelligence, Wuhan University, Wuhan, China,\\
 \textsuperscript{2}School of Computer Science, Wuhan University, Wuhan, China,\\
 \textsuperscript{3}School of Computer Science, Shanghai Jiao Tong University, Shanghai, China,\\
 \textsuperscript{4}Xiaomi Inc., Beijing, China
\\
\texttt{\{yanghq, zcli-charlie\}@whu.edu.cn, yaoyao27@sjtu.edu.cn,}\\
\texttt{\{qibaoyuan, liuguoming\}@xiaomi.com, zhaohai@cs.sjtu.edu.cn}\\
}

\begin{document}
\maketitle

\renewcommand{\thefootnote}
{\fnsymbol{footnote}} 
\footnotetext[1]{Corresponding author.} 

\input{pages/0-abstract}

\input{pages/1-introduction}
\input{pages/2-related-work}

\input{pages/3-xquant}

\input{pages/4-evaluation}

\input{pages/5-conclusion}

\input{pages/6-tmp}

\bibliography{custom}

\clearpage

\appendix

\input{pages/7-appendix}

\end{document}

%% file: pages/0-abstract.tex
\begin{abstract}

% Large Language Models (LLMs) have demonstrated remarkable capabilities across diverse natural language processing tasks.
% However, their extensive memory requirements stemming from KV cache growth, especially during long-text understanding and generation, pose significant challenges for real-world deployment in resource-constrained environments.
% Quantization, as a promising approach that preserves historical information while reducing memory consumption, has garnered significant attention and expectations.
% We present XQuant, a training-free and plug-and-play framework that pushes KV cache quantization to ultra-low equivalent bit-width.
% XQuant introduces two key improvements over existing quantization methods: a computationally negligible data-free calibration approach and cross-layer KV cache compression, enabling ultra-low equivalent bit-width.
% Extensive experiments on TruthfulQA and LongBench demonstrate that XQuant achieves lower equivalent bit-width (< 1.4 bits) across various large language models compared to KIVI-2bit and AsymKV-1.5bit baselines, while attaining superior performance metrics, establishing a better trade-off between model performance and compression ratio.

Large Language Models (LLMs) have demonstrated remarkable capabilities across diverse natural language processing tasks. However, their extensive memory requirements, particularly due to KV cache growth during long-text understanding and generation, present significant challenges for deployment in resource-constrained environments. Quantization has emerged as a promising solution to reduce memory consumption while preserving historical information. We propose XQuant, a training-free and plug-and-play framework that achieves ultra-low equivalent bit-width KV cache quantization. XQuant introduces two key innovations: a computationally negligible data-free calibration method and cross-layer KV cache compression, enabling quantization to sub-1.4 bits. Extensive experiments on TruthfulQA and LongBench demonstrate that XQuant outperforms state-of-the-art methods (e.g., KIVI-2bit and AsymKV-1.5bit) by achieving lower bit-width while maintaining superior performance, establishing a better trade-off between memory efficiency and model accuracy.
The source code is available at \url{https://github.com/brinenick511/XQuant}.

\end{abstract}

%% file: pages/1-introduction.tex
\section{Introduction}

The rapid advancement of Large Language Models (LLMs) has propelled significant progress in a wide array of natural language processing (NLP) applications, including code generation, search systems, and many others \cite{ouyang2023llm, sharma2024generative, ma2024comprehensive}. The exceptional performance of LLMs is primarily driven by their immense parameter scales, which enable them to excel across diverse tasks. However, this remarkable success comes with substantial costs: the computational and memory demands associated with deploying LLMs have increased exponentially due to increasing models parameters and growing input and output, posing a formidable bottleneck for practical deployment. In particular, GPU memory consumption has surged to levels that frequently surpass the capacities of current hardware infrastructures, making large-scale deployment increasingly challenging \cite{shi2024keep}.

% To alleviate this challenge, the Key-Value (KV) cache mechanism has been widely applied.
% By storing and reusing previously computed keys and values in the attention mechanism, the KV cache significantly reduces redundant computations and the GPU memory usage.
% But as the model sizes and the sequence lengths of model inputs and outputs continue to grow, the storage overhead of the KV cache itself continue to grow considerably as well.
% For example, a 30-billion-parameter language model with a batch size of 128 and a sequence length of 1024 can require up to 180 GB of memory solely for storing the KV cache \cite{zhang2023h2o}.
% This escalation has posed significant challenges for deploying LLMs in real-world environments with limited hardware resources.
% Consequently, exploring methods to compress KV caches without severely sacrificing model performance has become a highly valuable and critical research topic.

To mitigate this challenge, the Key-Value (KV) cache mechanism has been widely adopted \cite{yao2024sirllm,yang2024kvsharer,ainslie-etal-2023-gqa,kwon2023efficient}. The KV cache optimizes memory efficiency by storing and reusing previously computed keys and values in the attention mechanism, thereby reducing redundant computations and GPU memory usage. Despite its advantages, as model sizes and the input/output sequence lengths continue to grow, the storage overhead of the KV cache itself becomes increasingly significant \cite{shi2024keep}. For instance, a 30-billion-parameter language model with a batch size of 128 and a sequence length of 1024 may require up to 180 GB of memory solely for storing the KV cache \cite{zhang2023h2o}. 
Although the computational and memory requirements are reduced compared to not using it, such escalating demands still pose substantial challenges for deploying LLMs with constrained hardware resources.
% Such escalating demands pose substantial challenges for deploying LLMs in environments with constrained hardware resources. Consequently, compressing KV caches without significantly sacrificing model performance has emerged as a highly critical and valuable research topic.

% Prior works aim to solve this problem from different perspectives.
% Some previous studies \cite{sheng2023flexgen,hooper2024kvquant,Liu2024kivi,tao2024asymkv} map the floating-point KV cache (and even model weights) into discrete levels and store the quantized cache at lower precision only, but their performances may degrade under a extreme compression ratio around 2-bit.
% Some other works \cite{xiao2023streamingllm,zhang2023h2o,li2024snapkv,cai2024pyramidkv}
% focus on evicting unimportant tokens.

To address this problem, prior works have explored various strategies from different perspectives. Some studies \cite{sheng2023flexgen, hooper2024kvquant, Liu2024kivi, tao2024asymkv} focus on quantizing the floating-point KV cache (and, in some cases, model weights) to lower precision. However, these approaches often experience performance degradation under extreme compression ratios, particularly around 2-bit precision. Alternatively, other methods \cite{xiao2023streamingllm, zhang2023h2o, li2024snapkv, cai2024pyramidkv} aim to alleviate the storage burden by evicting unimportant tokens. These methods dynamically or statically identify and discard less critical tokens to reduce memory usage. Nevertheless, these methods inherently introduce information loss, resulting in reduced memory retention and severe forgetting issues, which can undermine the model's ability to maintain consistent performance on longer sequences.
% () also explore the possibility of reducing the number of heads or layers. 
% In addition to quantization and token eviction, some studies have also explored the possibility of reducing the number of heads or layers in the model \cite{yu2024effectively,yang2024laco}. 
% Existing approaches to KV cache quantization often prioritize computational efficiency but tend to sacrifice precision, particularly when operating under ultra-low-bit settings.
Existing KV cache quantization methods, due to inherent architectural constraints, fail to mitigate the severe performance degradation when operating under ultra-low-bit settings.

To address these limitations, this paper focuses on training-free KV cache quantization scenarios under extreme compression ratios and introduces \textbf{XQuant, a plug-and-play framework for ultra-low-bit KV cache quantization.} XQuant delivers two key improvements over existing quantization methods: 
\textbf{(1) Data-Free Calibration:} Traditional quantization methods often face significant limitations when mapping values to low-bit precision. Specifically, they tend to use the two endpoint values (e.g., 0 and 1 in 1-bit quantization) as representative values, which can result in substantial quantization errors, particularly under low bit-width settings. To address this issue, XQuant introduces a parameterized calibration scheme that allows for more fine-grained mapping of values. By adjusting the representative values to better reflect the actual data distribution, this method significantly reduces quantization errors and minimizes performance loss without the need for additional data. 
\textbf{(2) Cross-Layer KV Cache Compression:}
% We observe a high degree of similarity between quantized KV caches in adjacent layers. Leveraging this observation, we propose a cross-layer sharing strategy, where the quantized KV cache from one layer is shared across subsequent layers, thereby avoiding redundant computations. This simple yet effective strategy reduces the computational and memory overhead dramatically, while maintaining sufficient precision for downstream tasks.
We observe enhanced KV cache similarity between adjacent layers after quantization - a previously overlooked phenomenon.
% This enables effective cross-layer compression where one layer's quantized KV cache is shared across subsequent layers, significantly reducing computational and memory costs while preserving model performance.
This enables effective cross-layer compression, where the quantized KV cache of one layer is shared across subsequent layers, significantly reducing computational and memory costs. Meanwhile, a subset of layer-specific parameters is preserved to retain the unique characteristics of each layer, ensuring minimal loss of model performance.

To evaluate the effectiveness of XQuant, we conduct extensive experiments on a consumer-grade NVIDIA GeForce RTX 3090 GPU (24GB) across diverse datasets, including TruthfulQA \cite{TruthfulQA} and subsets of LongBench \cite{LongBench}. Experimental results demonstrate that XQuant achieves an equivalent bit-width of less than 1.4-bit across various LLMs, outperforming existing methods such as KIVI-2bit \cite{Liu2024kivi} and AsymKV-1.5bit \cite{tao2024asymkv}. Notably, XQuant achieves comparable performance to full-precision baselines while offering a significantly improved trade-off between model performance and compression ratio.

%% file: pages/2-related-work.tex
\section{Related Work}
Two mainstream approaches for addressing KV cache challenges are Quantization and Eviction methods \cite{shi2024keep}.

\textbf{Quantization} has emerged as a prominent technique for compressing large-scale models by mapping high-precision data to lower-precision formats (e.g., 16-bit, 8-bit, or even 4-bit integers). This significantly reduces memory footprints while maintaining acceptable levels of model performance. A substantial body of work focuses on quantizing model weights. AWQ \cite{lin2024awq} optimizes neural network weight quantization by dynamically adapting the bit-width based on the weights' significance. By retaining higher precision for more impactful weights and reducing precision for less critical ones, AWQ minimizes performance loss while achieving compression.
However, aggressive compression is constrained by "model hemorrhage" \cite{ma2025model}, a phenomenon identifying that models possess inherent robustness thresholds beyond which performance degrades sharply. This makes maintaining stability in the ultra-low-bit regime a critical challenge.
% However, this pursuit of aggressive compression is not without peril. The phenomenon of "model hemorrhage" highlights that models possess inherent robustness thresholds; once quantization pushes past these points, performance can degrade precipitously \cite{ma2025model}. This underscores the critical challenge of maintaining stability in the ultra-low-bit regime.

Another line of research concentrates on the quantization of the KV cache. KVQuant, introduced by \citet{hooper2024kvquant}, employs distinct quantization strategies for keys and values. It applies per-channel quantization to the keys—particularly before Rotary Positional Embeddings (RoPE)—and per-token quantization to the values, effectively managing outliers and minimizing RoPE-induced distortions. Similarly, MiKV \cite{DBLP:journals/corr/abs-2402-18096} introduces a mixed-precision KV-cache strategy that retains important KV pairs in high precision.
Concurrently, KIVI \cite{Liu2024kivi} develops a tuning-free 2-bit KV cache quantization scheme, where the key cache is quantized per-channel, and the value cache is quantized per-token. Building on this, AsymKV \cite{tao2024asymkv} further combines 1-bit and 2-bit representations through an asymmetric and layer-wise quantization configuration, achieving a better trade-off between precision and compression ratio.

In contrast, some works simultaneously quantize both the model weights and the attention cache. For example, FlexGen \cite{sheng2023flexgen} introduces a high-throughput inference framework that applies group-wise 4-bit quantization to compress both the model weights and KV cache. FlexGen divides tensors into small groups, computes the minimum and maximum values within each group, and performs asymmetric quantization. The resulting tensors are stored in 4-bit format and later dequantized to FP16 during computation, achieving a reduction in memory usage and I/O costs with minimal accuracy degradation.
Despite the advancements of these methods, significant performance degradation remains a challenge when quantizing KV cache activations to extremely low-precision levels, particularly below 2-bit. 

\textbf{Eviction} methods aim to discard unnecessary tokens during inference to reduce memory usage. StreamingLLM \cite{xiao2023streamingllm} identifies the phenomenon of attention sinks, where initial tokens are retained to stabilize attention computations. StreamingLLM combines these attention sinks with a sliding window of recent tokens to introduce a rolling KV cache, effectively balancing memory efficiency and model performance. Building on this, SirLLM \cite{yao2024sirllm} uses token entropy to preserve critical tokens' KV cache and incorporates a memory decay mechanism to enhance LLMs' long-term memory while maintaining short-term reasoning abilities.

Other methods, such as H2O \cite{zhang2023h2o} and SnapKV \cite{li2024snapkv}, dynamically identify and evict non-important tokens based on attention scores. PyramidKV \cite{cai2024pyramidkv, DBLP:conf/acl/YangHGHZ024} observes that attention scores are more sparse in higher layers and accordingly allocates different memory budgets across layers.
SpindleKV \cite{tang2025spindlekv} further develops a hybrid approach to balance reduction across layers, combining attention-based eviction in deep layers with a codebook-based replacement strategy for shallow layers.
However, most existing KV eviction methods depend on attention scores to identify non-important tokens, which limits their compatibility with common optimizations like FlashAttention \cite{dao2023flashattention2}, reducing their practical usability.

% \textbf{Inter-layer redundancy.} Beyond the above intra-layer redundancy in KV caches, some studies have also explored the inter-layer redundancy. Some prior works \cite{wu2024layer,sun2024yoco,brandon2024cla} investigate the potential of caching only partial layers of the KV cache, while other structural approaches like KV-Latent \cite{shi2025kvlatent} reduce the dimensionality of K and V vectors into a latent space, but all of them cannot be applied without additional training. We further clarify the key differences and highlight our contributions in Appendix \ref{sec:comparison}.

\textbf{Structural Approaches} modify the model's architecture, in contrast to post-hoc data compression. For instance, some methods cache only partial layers of the KV cache \cite{wu2024layer,sun2024yoco,brandon2024cla}, while KV-Latent \cite{shi2025kvlatent} reduces the dimensionality of K and V vectors. A key characteristic of these approaches is that they all require additional training, which contrasts with our plug-and-play framework.
We further clarify the key differences and highlight our contributions in Appendix \ref{sec:comparison}.

Compared to existing methods, we introduce XQuant with two key innovations: (1) A novel, simple yet effective data-free calibration method that achieves superior compression performance even under ultra-low-bit settings, eliminating the need for additional calibration data. (2) cross-layer KV cache compression that leverages previously overlooked quantization-enhanced layer similarities to achieve significant memory and computational savings. While prior work has studied layer representation similarities, our approach uniquely exploits the quantization-enhanced similarities to enable effective ultra-low-bit compression.

%% file: pages/3-xquant.tex
\section{XQuant}

\begin{figure*}
    \centering
    \includegraphics[width=1\linewidth]{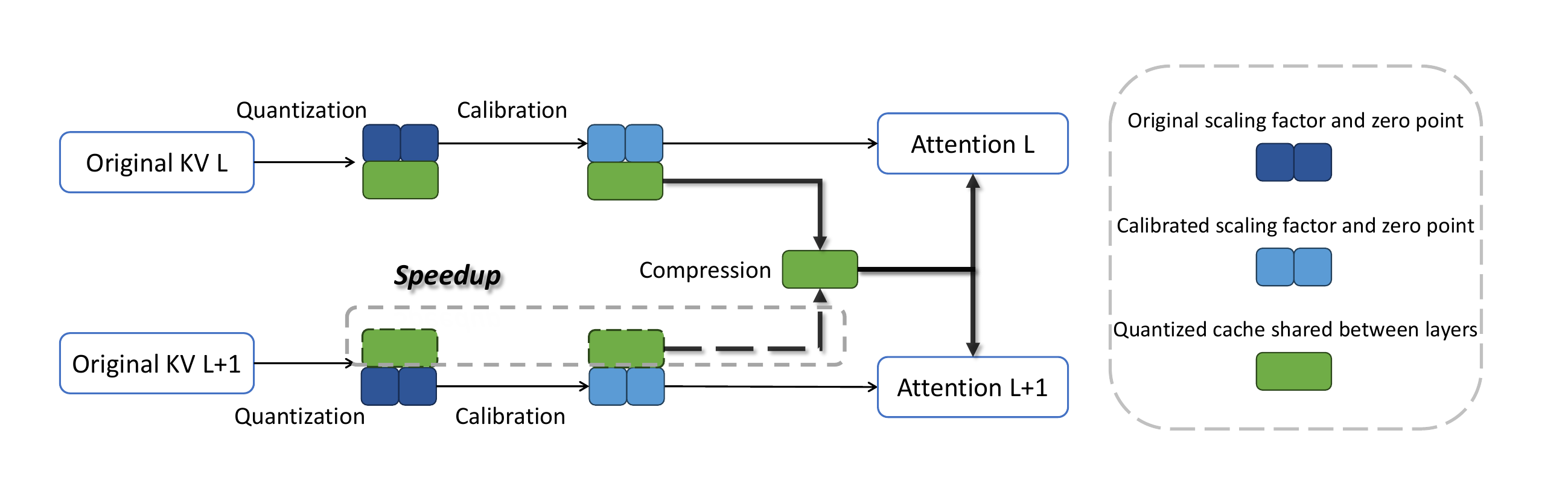}
    \caption{The illustration of XQuant workflow.
    % We group the KV cache layer-wise in pairs.
    XQuant partitions the KV cache into layer-wise pairs.
    For every higher layer in a pair, XQuant only computes and stores the scaling factors and zero-points during quantization phase, and then fetches the quantized cache from the lower layer during dequantization phase.
    }
    \label{fig:workflow}
\end{figure*}

% In this section, we introduce our XQuant with calibration (\sref{calibration}) and cross-layer compression (\sref{cross}) based on a classical KV cache quantization scheme (\sref{quant}).
% In this section, we first introduce the classical KV cache quantization scheme (\sref{quant}), which forms the foundation for our proposed approach. Building on this traditional framework, we present XQuant, which incorporates two key innovations: Calibration (\sref{calibration}) and Cross-Layer KV Cache Sharing Compression (\sref{cross}). These techniques address the limitations of classical methods by significantly reducing quantization errors and optimizing memory efficiency while maintaining high model performance.

In this section, we present XQuant, a novel quantization framework for efficient KV cache compression. As illustrated in \fref{workflow}, our framework introduces two key innovations: a data-free calibration technique that asymmetrically adjusts quantization parameters without additional calibration data, and a cross-layer KV cache compression mechanism that leverages the similarity of quantized caches between adjacent layers to effectively reduce both computational and memory overhead.

\subsection{Background}
\label{sec:quant}

To formalize KV cache quantization, we consider a group of floating-point keys or values \( \mathbf{X} \). The quantization process transforms \( \mathbf{X} \) into three components: a B-bit quantized cache \( \mathbf{X_Q} \), a zero-point \( z \), and a scaling factor \( s \) \cite{Liu2024kivi}:
\newline

\textbf{Quantization Phase:}
\begin{equation}
\label{eq:1}
    z=min(\mathbf{X}), s=\frac{max(\mathbf{X})-min(\mathbf{X})}{(2^B-1)}
\end{equation}
\begin{equation}
\label{eq:2}
    \mathbf{X_T}=(\mathbf{X}-z)/s, \mathbf{X_Q} = \lceil \mathbf{X_T} \rfloor
\end{equation}

\textbf{Dequantization Phase:}

\begin{equation}
    \mathbf{\hat{X}} = \mathbf{X_Q}*s+z
\end{equation}

\noindent where \(\mathbf{X^*}\) is the dequantized counterpart and \(\lceil \cdot \rfloor \) is the rounding function.
% \(\mathbf{X_T}\), the transformed matrix, is not cached but added in formula just for convenience.
\(\mathbf{X_T}\), the transformed matrix, is not explicitly cached but is introduced as an intermediate variable to facilitate subsequent mathematical derivations.

% Some prior work applies different configurations based on their observations, e.g., \cite{Liu2024kivi} focuses on the element distribution of KV cache and quantizes the key cache per-channel and the value cache per-token, and \cite{tao2024asymkv} applies different configurations to the key and value cache at the layer level. We basically follows these configurations for better performance.
% But these methods often face various degrees of performance degradation around 2-bit. To further compress the KV cache while maintaining adequate performance, we propose a novel and data-free calibration method based on our observations.

Building upon this framework, prior works introduce various configurations to enhance performance. For example, \citet{Liu2024kivi} focuses on the element-wise distribution within the KV cache, adopting per-channel quantization for the key cache and per-token quantization for the value cache. Similarly, \citet{tao2024asymkv} introduces layer-wise quantization configurations, employing asymmetric bit-widths for the key and value caches across different layers. While effective, these approaches often suffer from significant performance degradation under low-bit quantization settings, particularly around 2-bit precision. This limitation motivates the need for further advancements in KV cache compression techniques.

\subsection{Data-Free Calibration}
\label{sec:calibration}

Since existing quantization methods often experience significant performance degradation at 2-bit precision, achieving ultra-low-bit compression first requires bridging this performance gap.
In this section, we propose a data-free calibration method that effectively preserves model performance, enabling more aggressive compression ratios.

\begin{figure}
    \centering
    \includegraphics[width=1\linewidth]{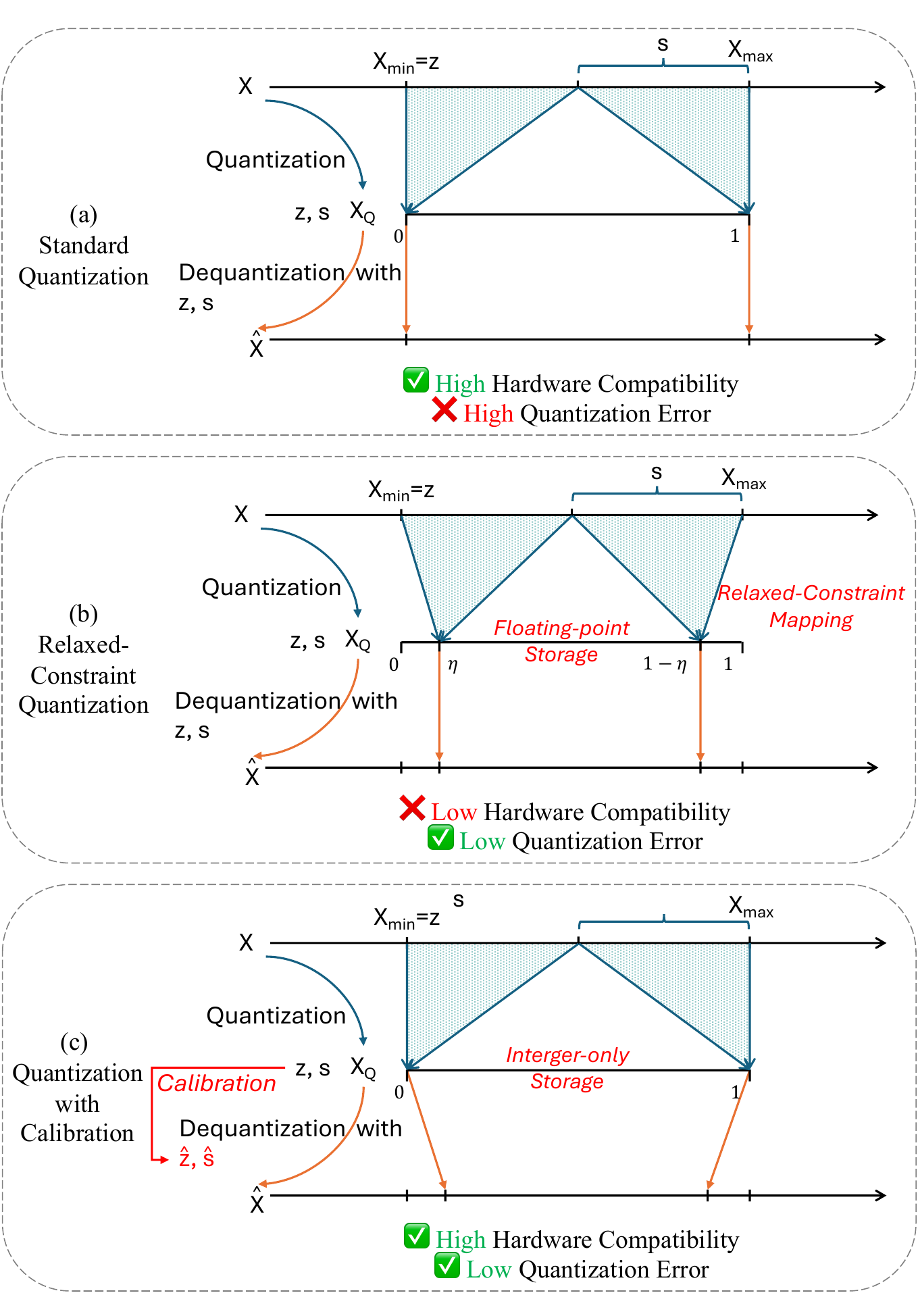}
    \caption{The illustration of the proposed data-free calibration method.
    }
    \label{fig:test}
\end{figure}

To analyze extreme quantization scenarios, we start with 1-bit quantization where each parameter is constrained to a binary state. Formally, the round-to-nearest operation \( \lceil \cdot \rfloor \) is defined as:

\begin{equation}
    \lceil e \rfloor =
    \begin{cases} 
    0 & \text{if } e \in [0, 0.5], \\
    1 & \text{if } e \in (0.5, 1].
    \end{cases}
\end{equation}

\noindent where \(e\) denotes an element of the transformed matrix. For any bit-width $B$, this rounding operation maps values to a discrete set within \([0,2^B-1]\), where each original value is assigned to its nearest representative in the quantized space.
As shown in Figure \ref{fig:test}(a), fixed representative values at endpoints (0 and 1) yield substantial quantization error for 1-bit quantization. We therefore introduce a relaxed-constraint mapping function that adaptively determines the quantization levels, formulated as:

\begin{equation}
\label{eq:5}
    f(e,\eta) =
    \begin{cases} 
    \eta & \text{if } e \in [0, 0.5], \\
    1-\eta & \text{if } e \in (0.5, 1].
    \end{cases}
\end{equation}

\noindent where \(\eta \in [0, 0.5] \) serves as a calibration parameter for determining quantization tendencies. Clearly, $f(e,0)$ is equivalent to the round-to-nearest function $\lceil e \rfloor$.
We extend this formulation to the general case of \(B\)-bit quantization and denote the corresponding parameter as \(\eta _B\).

We relax the constraint that quantized values must be integers and apply fake quantization as a preliminary experiment. Table \ref{tab:pre} shows that using this constraint-relaxed mapping function improves model performance, validating our proposed insight.

However, storing floating-point numbers as so-called quantized caches is impractical, as shown in Figure \ref{fig:test}(b).
To address the aforementioned problem, we establish an equivalent implementation, with the mathematical proof provided below.
We formalize the final data-free calibration approach as:

Consider a group of floating-point keys or values \(\mathbf{X} \in \mathbf{R}^{g}\), where \(g\) stands for the group size.
Note that \(\mathbf{X} \in [min(\mathbf{X}),max(\mathbf{X})]^{g} = [z,s*(2^B-1)+z]^{g}\), we can deduce:
\begin{equation}
    \mathbf{X_Q} \in [0,2^B-1]^{g}
\end{equation}
from \eref{1} and \eref{2}.
If we choose \(\eta*(2^B-1)\) and \((1-\eta)*(2^B-1)\) generalized from \eref{5} as two endpoints, it is equivalent to calibrate the zero-point and scaling factor to \(\hat{z}\) and \(\hat{s}\), and then dequantize with them.
Note that the dequantized matrix
\begin{equation}
    \mathbf{\hat{X}} = \mathbf{X_Q}*\hat{s}+\hat{z} \in [\hat{s}*0+\hat{z},\hat{s}*(2^B-1)+\hat{z}]^{g}
\end{equation}
and the corresponding interval given by two endpoints:
\begin{equation}
    [z+\eta s(2^B-1), z+s(2^B-1)(1-\eta)]
\end{equation}
By calculation we get the final operations for calibration:
\begin{equation}
    \hat{z} = z+\eta s(2^B-1), \hat{s}=(1-2\eta )s
\end{equation}

Since $\mathbf{X_T}=(\mathbf{X}-z)/s$, the reconstruction loss $MSE(\mathbf{X},\mathbf{\hat{X}}) = s^2 \cdot MSE(\mathbf{X_T},f(\mathbf{X_T},\eta))$.
For analytical tractability, particularly for 1-bit quantization within small group sizes, we can assume that $\mathbf{X_T} \sim U(0,1)$. Thus the expected MSE in the transformed space can be formulated as:
\begin{align*}
&MSE(\mathbf{X_T},f(\mathbf{X_T},\eta)) \\
& = E[(X_T - f(X_T, \eta))^2] \\
& = \int_0^{0.5} (x - \eta)^2 dx + \int_{0.5}^1 (x - (1-\eta))^2 dx \\
& = \eta^2 - \frac{1}{2}\eta + \frac{1}{12}
\end{align*}

Since the standard quantization scheme is equivalent to setting $\eta=0$, this result confirms that any value of $\eta \in (0,1/2)$ will strictly reduce the theoretical reconstruction error.

As shown in Figure \ref{fig:test}(c), we propose the improved quantization scheme with this data-free calibration as follows:
\newline

\textbf{Quantization Phase with Calibration:}

\begin{equation}
    z=min(\mathbf{X}), s=\frac{max(\mathbf{X})-min(\mathbf{X})}{(2^B-1)}
\end{equation}
\begin{equation}
    \mathbf{X_T}=(\mathbf{X}-z)/s, \mathbf{X_Q} = \lceil \mathbf{X_T} \rfloor
\end{equation}
% \begin{equation}
%     \hat{z}=z+s*\eta, \hat{s}=s*(1-2*\eta)
% \end{equation}
\begin{equation}
\label{eq:12}
    % \hat{z}=z+s*\eta * (2^B-1), \hat{s}=s*(1-2*\eta)
    \hat{z} = z+\eta s(2^B-1), \hat{s}=(1-2\eta )s
\end{equation}

\textbf{Dequantization Phase with Calibration:}

\begin{equation}
    \mathbf{\hat{X}} = \mathbf{X_Q}*\hat{s}+\hat{z}
\end{equation}

\begin{figure}
    \centering
    \includegraphics[width=0.95\linewidth]{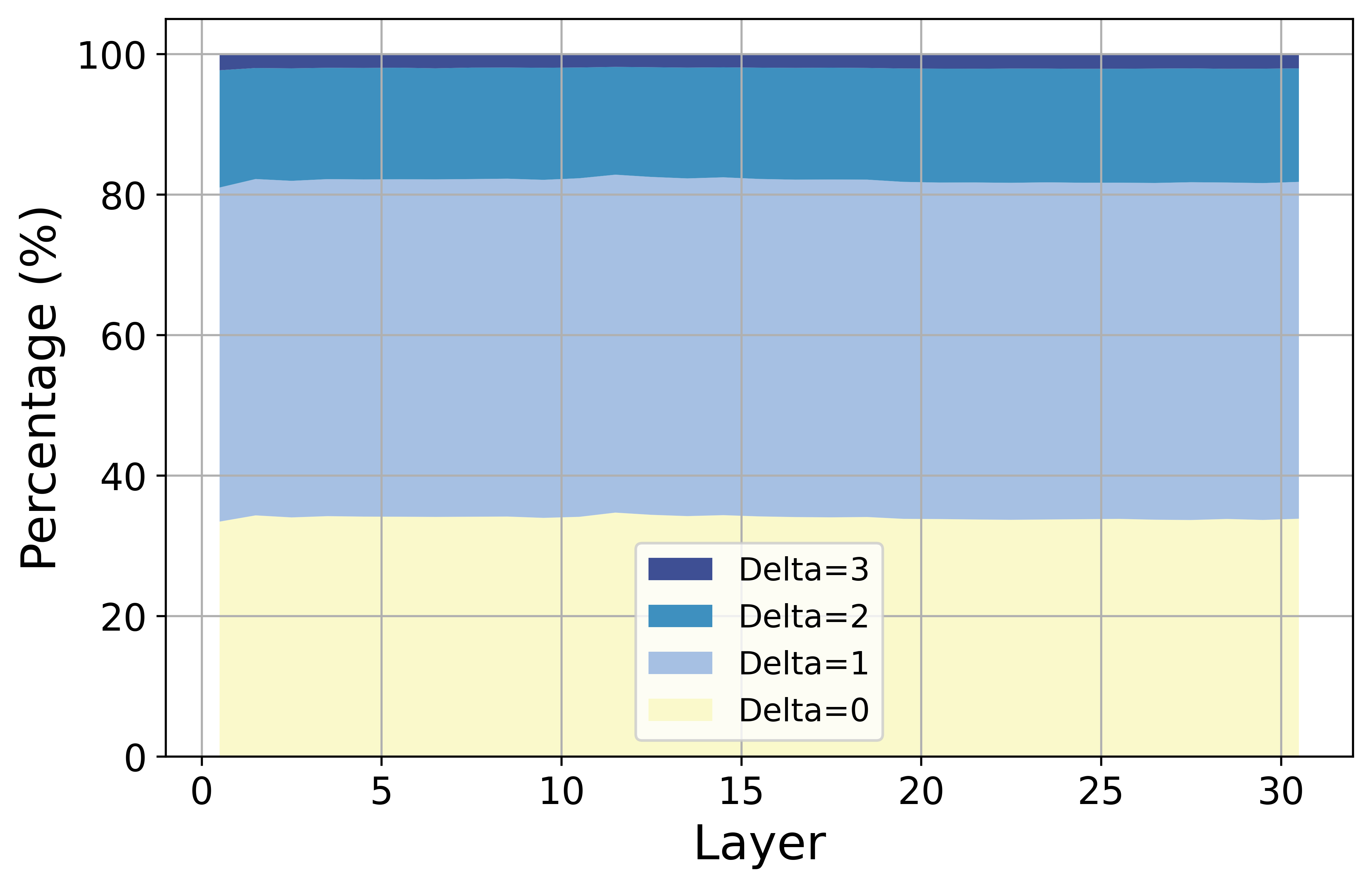}
    \caption{Layer-wise analysis of absolute differences between adjacent layers in quantized KV Cache matrices.
    Here, delta represents the absolute difference of quantized values between consecutive layers.
    % The results indicate a dominant presence of small differences (0 and 1), demonstrating a high degree of similarity between adjacent layers.
    }
    \label{fig:delta}
\end{figure}

\subsection{Cross-Layer Compression}
\label{sec:cross}

% On LongBench \cite{bai2023longbench} tasks, the key cache is nearly impossible to be quantized under 2-bit in former asymmetrical quantization scheme \cite{tao2024asymkv}. Our reproduction experiments show that only quantizing the higher 8 layers of the key cache into 1-bit (reproduced asym-24/0) still causes a severe performance degredation. To push low-bit KV cache quantization to the limit, we proposes a novel cross-layer compression method to further compress a value layer into equivalently 0.5-bit.
% Following the findings of \citet{tao2024asymkv}, our reproduction experiments on LongBench tasks \cite{bai2023longbench} reveal that key cache quantization under the 2-bit asymmetrical quantization scheme is nearly infeasible. Even when limiting the quantization of the higher 8 layers of the key cache to 1-bit (reproduced as asym-24/0), the model still suffers from severe performance degradation. To push low-bit KV cache quantization further, we propose a novel cross-layer compression method that effectively compresses the value cache to an equivalent 0.5-bit representation.

\subsubsection{Motivation}

Building upon \citet{tao2024asymkv}'s investigation of ultra-low-bit KV cache asymmetric quantization, our reproduction experiments on LongBench \cite{bai2023longbench} with Mistral \cite{jiang2023mistral} demonstrate severe limitations of existing approaches, as shown in Table \ref{tab:obs}.

% Our investigation revealed that 1-bit asymmetric quantization of the Key cache is practically infeasible. Even when limiting 1-bit quantization to the top 8 layers (AsymKV-24/32), we observed significant performance degradation.
% Given the challenges and limitations of further quantizing the key cache, we turn our attention to exploring cross-layer compression methods.
% These methods offer a promising avenue for achieving similar ultra-low-bit quantization while mitigating the performance trade-offs associated with direct quantization.

We found that 1-bit asymmetric quantization of the key cache is practically infeasible. Even when restricting 1-bit quantization to the top 8 layers (AsymKV-24/32), significant performance degradation occurs. Given the limitations of further key cache quantization, we turn to cross-layer compression techniques as a viable alternative to achieve comparable ultra-low-bit quantization without compromising performance.

\subsubsection{Analysis on Quantized KV Cache}
% To explore the feasibility of further compression, we first examine the similarity of quantized KV caches between adjacent layers. Specifically, we investigate whether adjacent layers exhibit significant redundancy, which would provide greater opportunities for aggressive quantization. To this end, we adopt the KIVI-2 framework \cite{Liu2024kivi} and perform inference on randomly selected data from LongBench \cite{bai2023longbench} using the Mistral-7B-Instruct-v0.2 model \cite{jiang2023mistral}.

To enable cross-layer compression, we first analyze the characteristics of quantized KV caches by examining inter-layer similarities. We hypothesize that significant redundancy between adjacent layers could create opportunities for more aggressive compression. Using the KIVI-2 framework \cite{Liu2024kivi}, we conduct preliminary experiments on the Mistral-7B-Instruct-v0.2 model \cite{jiang2023mistral} with random samples from LongBench \cite{bai2023longbench}.
% We collect the KV cache from both the prefilling phase and the decoding phase, and calculate the element-wise absolute differences between the Value cache tensors of adjacent layers for all attention heads. Specifically, for each pair of adjacent layers, we compute the element-wise differences of Value cache and analyze the distribution of the resulting element values.
% Our analysis encompasses KV caches from both prefilling and decoding phases. For each pair of adjacent layers, we compute element-wise absolute differences between value cache tensors across all attention heads, allowing us to characterize the distribution of inter-layer variations.

Under the 2-bit quantization scheme in KIVI-2, quantized cache values are restricted to \{0, 1, 2, 3\}, naturally constraining element-wise absolute differences to the same range. Our analysis, illustrated in Figure \ref{fig:delta}, reveals a striking pattern: over 80\% of positions between adjacent layers exhibit minimal differences (0 or 1), while extreme differences (3) occur in less than 5\% of positions. This pattern becomes even more pronounced in the 1-bit scenario, where mapping \{0,1\} to 0 and \{2,3\} to 1 maintains identical values in over 80\% of positions between adjacent layers. These empirical findings demonstrate substantial redundancy in quantized KV caches between adjacent layers, suggesting significant potential for further compression. 
% Our observations align with concurrent research by \cite{liu2024minicache}, who identified similar patterns using angular distance metrics, particularly in middle-to-deep layers.

% Under the classical 2-bit quantization scheme, the quantized cache values are constrained to the set \{0, 1, 2, 3\}, meaning the element-wise absolute differences are also limited to this range. The results are shown in \fref{delta}. We can see from the figure that over 80\% of the positions between adjacent layers have a difference of 0 or 1, while extreme differences of 3 occur in less than 5\% of the positions. Furthermore, in the 1-bit quantization scenario, where the element-wise differences \{0,1\} and \{2,3\} in 2-bit quantization correspond to 0 and 1, respectively, over 80\% of the positions remain identical between adjacent layers.

% This analysis provides strong evidence of the high degree of similarity between the quantized KV caches of adjacent layers, validating the potential for further compression. These findings are consistent with concurrent work by \citet{liu2024minicache}, which shows that KV caches in the middle-to-deep layers exhibit significant similarity between adjacent layers based on angular distance.

\subsubsection{Compression Algorithm}
Leveraging these insights into inter-layer similarities, we propose a novel cross-layer compression method that decomposes KV caches into two components: shared quantized caches and layer-specific parameters. Specifically, adjacent layers share a common set of quantized value caches ($\mathbf{X_Q}$), while maintaining their individual scaling factors and zero-points for dequantization.
This decomposition enables efficient compression by allowing each layer to reuse the merged cache from its group, while preserving the layer-specific characteristic through its unique quantization parameters, namely zero-points and scaling factors.
% The approach significantly reduces both memory footprint and computational overhead while maintaining model performance.
% Motivated by this observation, we propose a method to share the quantized value cache \(\mathbf{X_Q}\) between adjacent layers. Specifically, we group layers into pairs, where the higher layer reuses the quantized cache of its adjacent lower layer, applying only its own scaling factors and zero-points during the dequantization phase. This approach drastically reduces both memory overhead and computational cost while maintaining model performance. The workflow of our method is illustrated in \fref{workflow}.

% To analyze the impact of group size and the compression method, we conduct detailed ablation studies, as discussed in \sref{ablation}. Based on these experiments, we fix the group size at 2 and observe optimal performance with a degraded weighted average when \(\gamma=1\). By sharing quantized caches in layer pairs, we achieve a significant reduction in memory usage while preserving the accuracy of the model.
In the implementation, for a model with $L$ layers, we organize the layers into groups of size $G$. Within each group, KV caches are compressed using weighted averaging, where each layer $l$ ($0 \leq l \leq L$) is assigned a weight $\gamma_l$, subject to the constraint $\sum \gamma_l = 1$.

Formally, for every layer \(l\) in a group \(G\), the quantization workflow with cross-layer compression and calibration is utilized as follows:

\textbf{Quantization Phase with Cross-Layer Compression and Calibration:}

% for \(k = l, l+1,...\)
% \begin{equation}
%     z^l=min(\mathbf{X}^l)+\frac{(max(\mathbf{X}^l)-min(\mathbf{X}^l))*\eta_B}{2^B-1} 
% \end{equation}
% \begin{equation}
%     s^l=\frac{(max(\mathbf{X}^l)-min(\mathbf{X}^l)) * (1-2*\eta_B)}{2^B-1} 
% \end{equation}
% \begin{equation}
%     \mathbf{X_Q} = \lceil (\mathbf{X}^l-z^l)/s^l \rfloor
% \end{equation}
% % \begin{equation}
% %     z^{l+1}=min(\mathbf{X}^{l+1})+  \frac{(max(\mathbf{X}^{l+1})-min(\mathbf{X}^{l+1}))*\eta_B}{2^B-1} 
% % \end{equation}
% \begin{align}
%     z^{l+1} &= \min(\mathbf{X}^{l+1}) + \\
%     & \qquad \frac{\big( \max(\mathbf{X}^{l+1}) - \min(\mathbf{X}^{l+1}) \big) \cdot \eta_B}
%     {2^B - 1}
% \end{align}

% \begin{align}
%     s^{l+1} =\frac{(max(\mathbf{X}^{l+1})-min(\mathbf{X}^{l+1})) * (1-2*\eta_B)}{2^B-1} 
% \end{align}

% \begin{align}
%     z=min(\mathbf{X}) \\
%     s=\frac{max(\mathbf{X})-min(\mathbf{X})}{(2^B-1)} \\
%     z^l = &\min(\mathbf{X}^l) + \frac{\big( \max(\mathbf{X}^l) - \min(\mathbf{X}^l) \big) \cdot \eta_B}
%     {2^B - 1} \\
%     s^l =& \frac{\big( \max(\mathbf{X}^l) - \min(\mathbf{X}^l) \big)\cdot (1 - 2 \cdot \eta_B)}
%     {2^B - 1} 
% \end{align}

% \begin{align}
%     z^{l+1} =& \min(\mathbf{X}^{l+1}) + \nonumber \\
%     & \qquad  \frac{\big( \max(\mathbf{X}^{l+1}) - \min(\mathbf{X}^{l+1}) \big) \cdot \eta_B}
%     {2^B - 1} \\
%     s^{l+1} = &(1 - 2 \cdot \eta_B) \cdot \nonumber\\
%    & \qquad \frac{\big( \max(\mathbf{X}^{l+1}) - \min(\mathbf{X}^{l+1}) \big)}
%     {2^B - 1} 
% \end{align}

\begin{equation*}
    % for \ i = 0, 1, \dots, G-1
    \forall l \in \mathbf{G},
\end{equation*}
\begin{equation*}
    z_{l}=min(\mathbf{X}_{l}),  \
% \end{equation*}
% \begin{equation*}
    s_{l}=\frac{max(\mathbf{X}_{l})-min(\mathbf{X}_{l})}{(2^B-1)}
\end{equation*}
\begin{equation*}
    \hat{z}_{l} = z_{l}+\eta s_{l} (2^B-1), \
% \end{equation*}
% \begin{equation*}
    \hat{s}_{l}=(1-2\eta )s_{l}
\end{equation*}

\begin{equation*}
\label{eq:45}
        \mathbf{X_Q} = \sum_{l \in \mathbf{G}}\gamma_{l} \bigg\lceil \frac{\mathbf{X}_{l} - z_{l}}{s_{l}} \bigg\rfloor
\end{equation*}

\textbf{Dequantization Phase with Cross-Layer Compression and Calibration:}
% \begin{equation}
%     \mathbf{\hat{X}}^l = \mathbf{X_Q}*\hat{s}^l+\hat{z}^l
% \end{equation}
\begin{equation*}
    \mathbf{\hat{X}}_{l} = \mathbf{X_Q}*\hat{s}_{l}+\hat{z}_{l}
\end{equation*}

We present the pseudo code for the whole workflow as shown in Appendix \ref{sec:algorithms}.

\begin{table}
  \centering
  \setlength{\extrarowheight}{1pt}
  \scalebox{0.9}{
    \begin{tabular}{lllc}
    \hline
    \textbf{Model} & 
    \textbf{Method} &
    \textbf{Bit-width} &
    \textbf{TruthfulQA}
    \\
    \hline
    \multirow{4}{*}{Mistral-7b} 
    & Full Cache & 16
    &  32.09 \\ \cline{2-4}
    & KIVI & 2
    & 32.17 \\
    & AsymKV & 1.5
    & 32.80 \\
    & XQuant & 1.38
    &\textbf{34.93} \\
    \hline
    \multirow{4}{*}{Llama2-7b} 
    & Full Cache & 16
    &30.77 \\ \cline{2-4}
    & KIVI & 2
    &33.92 \\
    & AsymKV & 1.5
    &33.84\\
    & XQuant & 1.4
    &\textbf{34.22}\\
    \hline
  \end{tabular}
  }
  \caption{\label{tab:eval}
    Evaluation on TruthfulQA task with normal context length.
  }
\end{table}

\begin{table*}
  \centering
  \small
  \setlength{\extrarowheight}{3pt}
  \resizebox{\textwidth}{!}{
  \scalebox{0.75}{
      \begin{tabular}{lllcc ccccc c}
        \hline
        \textbf{Model} & 
        \textbf{Method} &
        \textbf{Bit-width} &
        \textbf{HQA} &
        \textbf{2Wiki} &
        \textbf{MSQ} &
        \textbf{TREC} &
        \textbf{TQA} &
        \textbf{SAMS} &
        \textbf{PC} &
        \textbf{Avg}
        \\
        \hline
        \multirow{5}{*}{Mistral-7b-ins} 
        & Full Cache & 16
        & 43.02 &	27.10 &	18.78 &	 71.00 &	86.23&42.75& 2.75 &41.66 \\
        & PyramidInfer & /
        &35.08 	&23.92 	&16.90 	&62.00 &	85.06 	&41.45 &1.04&32.55\\
        \cline{2-11}
        & KIVI & 2
        & {\ul 41.96} 	&{\ul 26.08} &\textbf{18.13} &	{\ul 71.00} &	{\ul 86.00} &	{\ul 43.70} &	2.78 & {\ul 41.38} \\
        & AsymKV & 1.5
        &37.17 	&22.77 	&15.76 	& 70.50 	&\textbf{86.25} 	&43.44 &{\ul 3.16}& 39.86\\
        & XQuant & 1.38
        &\textbf{42.90}& \textbf{26.65} &	{\ul 17.44} 	&\textbf{71.50}& 	84.50 	&\textbf{45.18} &\textbf{5.71} &\textbf{41.98}\\
        \hline
        \multirow{5}{*}{Llama2-7b-chat} 
        & Full Cache & 16
        &30.09 &26.48 &	9.98 &	63.00 &	84.19 &	41.22 	&4.50 &37.07 \\
        & PyramidInfer & /
        &29.14 	&24.53 &	7.49 &	54.00 &	81.79 	&40.71 &	4.00 &34.52\\
        \cline{2-11}
        & KIVI & 2
        &{\ul 29.10} 	&{\ul 25.12}	&\textbf{9.86} &\textbf{63.00} &\textbf{84.98} 	&\textbf{40.18}& \textbf{4.00} & \textbf{36.61} \\
        & AsymKV & 1.5
        &27.75 &	24.82& 	8.45 &	62.00 	&84.21 	&41.22 &	{\ul 2.75} & 35.89\\
        & XQuant & 1.4
        &\textbf{29.21} &	\textbf{25.56} &	{\ul 9.69} &{\ul 62.50} 	&{\ul 84.57} 	&{\ul 40.01} 	&\textbf{4.00} &{\ul 36.51}\\
        \hline
      \end{tabular}
      }
  }
  \caption{\label{tab:longbench}
    Evaluation of different KV cache compression methods on LongBench tasks.
  }
  \vspace{-10pt}
\end{table*}

\subsubsection{Speedup through Cross-layer Compression}
\label{sec:speedup}

While our previous discussion introduced weighted averaging with the weight $\gamma$ for compressing $\mathbf{X_Q}$ within a group, we can further optimize the computation by setting $\gamma_{k} = 1$ for a chosen dominant layer $k$, which consequently forces all other $\gamma$ values within the group to zero.
% In this simplified configuration, $X_Q$ is derived solely from the KV cache of the chosen layer in each group, substantially reducing computational overhead.
In this accelerated configuration, each subordinate layer only needs to compute and store its own scaling factors and zero-points, significantly reducing computational overhead.
Specifically, 
\begin{equation*}
        \mathbf{X_Q} = \bigg\lceil \frac{\mathbf{X}_{k} - z_{k}}{s_{k}} \bigg\rfloor \\
\end{equation*}

As illustrated in \fref{workflow}, this optimization eliminates the computations shown in the dashed line, effectively streamlining the process.
Experimental results show that selecting the first layer within the group as the dominant layer yields optimal performance, as demonstrated in Table \ref{tab:gamma} and Table \ref{tab:group}.

%% file: pages/4-evaluation.tex
\section{Evaluation}

\subsection{Experimental Setup}

\paragraph{Models.}
We evaluate our XQuant on Llama-2-7b / Llama-2-7b-chat \cite{touvron2023llama} and Mistral-7B-v0.3 / Mistral-7B-instruct-v0.2 \cite{jiang2023mistral}.
% All the experiments are based on Huggingface() and PyTorch.

\paragraph{Tasks.}
For the normal context length task, we choose TruthfulQA (BLEU score) from LM-Eval \cite{gao2021eval}. We also select several subsets from LongBench \cite{bai2023longbench} for the long context length tasks, including HotpotQA (F1 score), 2WikiMultihopQA (F1 score), MuSiQue (F1 score), TREC (classification accuracy), TriviaQA (F1 score), SAMSum (Rouge-L) and PassageCount (Exact match accuracy). MultiFieldQA-Zh (F1 score) is selected for some ablation studies as well.

\paragraph{Baselines and Implementations.}
We compare our framework with previous works, including original 16-bit floating implementation, KIVI-2 \cite{Liu2024kivi} and AsymKV \cite{tao2024asymkv}. All relevant configurations adhere as in KIVI, i.e., quantizing key cache per-channel and value cache per-token, and with a group size of 32 and a residual length of 128.
We reproduce AsymKV based on the official implementation of KIVI, with a typical configuration (AsymKV-32/0) selected from the original paper, i.d., quantizing all the key cache into 2-bit and value cache into 1-bit, which corresponds to an equivalent bit-width of 1.5.

% We also choose a token eviction method \cite{yang2024pyramidinfer} for comparison on LongBench tasks as well, with a 40\% KV cache setting.

A token eviction method~\cite{yang2024pyramidinfer}, configured with a 40\% KV cache budget, is also included as a baseline for the LongBench tasks.

We set the maximum sequence length to 30000 for the Mistral model to conduct our experiments with a single NVIDIA GeForce RTX 3090 GPU (24GB), and 8192 for the Llama model as default.
We do not consider SLERP \cite{shoemake1985slerp,liu2024minicache} because of the incompatibility between rescale-recover operations and quantized cache.

\subsection{Performance Comparison}

\paragraph{LM-Eval Results.}
% The performance comparison between different compression methods is summarized in \tref{eval}. We apply XQuant, KIVI and AsymKV to Mistral-7b-v0.3 and Llama2-7b.
% As shown in \tref{eval}, with the lowest precision level, our XQuant surpass all other compression methods in TruthfulQA dataset while achieve a comparable performance in CoQA dataset with the lowest-precision quantization.
Table \ref{tab:eval} presents the evaluation of different quantization methods on the TruthfulQA task with a standard context length. XQuant not only achieves competitive performance but surpasses the full cache baseline, with a TruthfulQA score of 34.93 on Mistral-7b and 34.22 on Llama2-7b, outperforming all other methods at significantly lower bit-widths.
These results highlight that XQuant provides superior performance in conventional context length settings.

\paragraph{LongBench Results.}
% We benchmark our XQuant in the LongBench datasets with Mistral-7b-instruct-v0.2 and Llama-2-7b-chat.
% As shown in \tref{longbench}, our XQuant outperforms KIVI-2bit on average. Even under the lowest precision level, XQuant achieve comparable performance with 16-bit full cache baseline in most datasets.

We evaluate XQuant on the LongBench benchmark using two widely adopted models: Mistral-7b-Instruct-v0.2 and Llama-2-7b-chat. As shown in Table~\ref{tab:longbench}, XQuant achieves significant improvements over other KV cache compression methods, particularly under ultra-low-bit settings.

% XQuant retains performance comparable to the full cache baseline across datasets such as TREC (71.50 vs. 71.00).
In all datasets of LongBench, XQuant achieves performance comparable to the full cache baseline while reducing bit-width by 31\% compared to KIVI-2bit.
Notably, XQuant achieves an average score of 41.98 for Mistral, surpassing KIVI-2bit while maintaining a significantly lower bit-width of 1.38.
Moreover, XQuant outperforms AsymKV on nearly all datasets while simultaneously reducing bit-width by 8\% relative to AsymKV.
Additionally, compared to PyramidInfer, which sacrifices precision to reduce storage overhead, XQuant demonstrates clear advantages in maintaining high accuracy across tasks while achieving lower bit-width.
% XQuant consistently outperforms baselines across most tasks, achieving an average score of 36.44. 

\subsection{Ablation and Analysis}
\label{sec:ablation}
In this section, we conduct ablation studies in some randomly selected lightweight LongBench subsets.

\begin{table}
  \centering
  \small
  \setlength{\extrarowheight}{2pt}
\begin{tabular}{llccc}
\hline
\textbf{Method}         & \textbf{Bit-width}     & \textbf{\( \eta _1\)} & \textbf{\( \eta _2\)} & \textbf{MFQA-Zh} \\ \hline
Full Cache              & 16                     & /           & /           & 48.26            \\
KIVI                    & 2                      & /           & 0           & 42.27            \\
AsymKV                  & 1.5                    & 0           & 0           & 36.30            \\
\hline

\multirow{4}{*}{XQuant} & \multirow{4}{*}{1.375} & 0           & 0           & 37.20            \\
                        &                        & 0           & 0.05        & \textbf{40.32}           \\
                        &                        & 0.2         & 0           & \textbf{41.98}            \\
                        &                        & 0.2         & 0.05        & \textbf{44.20} 
                        \\
\hline
\end{tabular}
  \caption{\label{tab:eta}
    % The comparison using different quantization methods with and without our calibration method in MultiFieldQA-Zh tasks from LongBench.
    Ablation study on the effect of data-free calibration in XQuant on the MultiFieldQA-Zh benchmark from LongBench.
    % \e{i} = 0 means the i-bit quantization is deployed without calibration. The AsymKV quantizaiton is deployed with the default 1.5-bit configuration consistent with in other experiments in this paper, which means using 2-bit quantization for key cache and 1-bit quantization for value cache.
  }
\end{table}

\paragraph{Calibration Parameter.}

% Table \ref{tab:eta} shows the ablation study on the effect of data-free calibration in XQuant on the MultiFieldQA-Zh benchmark.
% The results demonstrate that incorporating calibration (where\( \eta != 0 \)) significantly enhances the performance of XQuant, narrowing the gap with the full cache baseline.

Table \ref{tab:eta} presents an ablation study on the impact of data-free calibration in XQuant on the MultiFieldQA-Zh benchmark. The results indicate that applying calibration (\(\eta_1 \neq 0\) or \(\eta_2 \neq 0\)) significantly improves XQuant's performance, reducing the performance gap with the full cache baseline.

% As described in \sref{calibration}, there are two parameters in our data-free calibration method, namely \e{1} for 1-bit quantization and \e{2} for 2-bit.
% We set \e{1} = 0 or 0.2 and \e{2} = 0 or 0.05 and deploy our calibration method on default KIVI-2bit and AsymKV-1.5bit to demonstrate its effectiveness.
% Note that \e{} = 0 is equal to not deploying our calibration method.
% As shown in \tref{eta}, these two quantization frameworks can perform better with our lightweight calibration method.

\vspace{10pt}
\begin{table}
  \centering
  \small
  \begin{tabular}{llcc}
    \hline
    \textbf{Method} &
    \textbf{Bit-width} &
    \textbf{\(\gamma_0\)} &
    \textbf{MuSiQue} \\
    \hline
    Full Cache & 16 & / & 18.78 \\
    KIVI & 2 & / & 18.13 \\
    \hline
    Flooring & 1.63 & / & 16.79 \\
    Ceiling & 1.63 & / &16.36 \\
    % Stochastic & 1.63 & / &17.65 \\
    \hline
    \multirow{6}{*}{Weighted Average}
    & 1.63 & [0,1/6) & 12.20 \\
    & 1.63 & (1/6,1/4) & 14.05 \\
    & 1.63 & (1/4,1/2) & 16.84 \\
    & 1.63 & (1/2,3/4) & {\ul 17.32}\\
    & 1.63 & (3/4,5/6) & \textbf{17.60} \\
    & 1.63 & (5/6,1] & {\ul 17.32} \\
    \hline
  \end{tabular}
  
  \caption{\label{tab:gamma}
    The comparison between different cross-layer compression method with group size $G = 2$, where \(\gamma_0, \gamma_1 \) stands for the coefficient in the weighted average \((\gamma_1 + \gamma_0= 1 )\).
    % We only compress the higher 24 of layers in the quantized value cache based on KIVI-2bit. Results show that the degraded compression method can still achieve a competitive performance when \( \gamma \in (5/6,1]\), which only requires the minimal computational cost.
  }
  % \vspace{-15pt}/
\end{table}

\paragraph{Cross-Layer Compression Method.}

% Rounding operation, such as flooring or ceiling, is an critical part when averaging quantized caches.
We further explore the weighted average with a group size \(G=2\) and coefficients \(\gamma_0, \gamma_1 = 1-\gamma_0\), where \(\gamma_0\) falls into six intervals derived in Appendix \ref{app:int}. Notably, when \(\gamma_0 \in [0,1/6)\) or \(\gamma_0 \in (5/6,1]\), the operation is optimized to directly sharing the quantized cache.
We evaluate KIVI-2 on Mistral-7B-Instruct-v0.2 without our proposed calibration methods starting from the 8-th layer.
As summarized in \tref{gamma}, the accelerated compression methods (\(\gamma_0 \in [0,1/6) \cup (5/6,1]\)) avoid redundant operations seen in the workflow of \citealp{Liu2024kivi}, which rounds quantized integers into floating-point numbers.
% These methods strike a better balance between efficiency and memory usage.
As shown in Table~\ref{tab:gamma}, the accelerated compression operation demonstrates its effectiveness in maintaining sufficient information for model performance, particularly when \(\gamma_0 \in (5/6, 1]\). This configuration effectively allows odd-numbered layers to reuse the quantized cache from the preceding even-numbered layers without requiring additional quantization or storage overhead for odd-numbered layers.

We adopt this accelerated compression strategy across all experiments due to its favorable balance between computational efficiency and information preservation.

\begin{table}
  \centering
  % \small
  \setlength{\extrarowheight}{2pt}
    \scalebox{0.85}{
  \begin{tabular}{llllcc}
\hline
\textbf{Method}          & \textbf{Bit-width}     & \textbf{\(G\)}      & \textbf{\(k\)} & \multicolumn{1}{l}{\textbf{MSQ}} & \multicolumn{1}{l}{\textbf{MFQA-Zh}} \\ \hline
Full Cache               & 16                     & /    & / & { 18.78}     & { 48.26}         \\
KIVI                     & 2                      & /    & / & { 18.13}     & { 42.27}         \\ \hline
 &  &                     & 0 & { \textbf{17.32}} & { \textbf{37.44}} \\
 &  & \multirow{-2}{*}{2} & 1 & { 12.20}          & { 20.48}          \\ \cline{3-6} 
 &  &                     & 0 & { 14.92}          & { 17.53}          \\
 &  &                     & 1 & { 16.97}          & { 37.37}          \\
 &  & \multirow{-3}{*}{3} & 2 & { 13.21}          & { 20.80}          \\ \cline{3-6} 
 &  &                     & 0 & { 14.82}          & { 23.53}          \\
 &  &                     & 1 & { 12.44}          & { 18.68}          \\
 &  &                     & 2 & { 16.12}          & { 35.48}          \\
\multirow{-9}{*}{XQuant} & \multirow{-9}{*}{1.63} & \multirow{-4}{*}{4} & 3                & { 15.39}     & { 20.32}         \\ \hline
\end{tabular}
}
  \caption{\label{tab:group}
    The comparison of different group sizes \(G\) and selection indices \(k\) within each group, where XQuant is employed without the calibration step for a clearer analysis.
    % , where \(id\) means we only store and share the quantized cache in the idx-th layer form each group. The results show that this compression method performs well especially with a group\_size of 2 and a idx of 0.
  }
  % \vspace{-10pt}
\end{table}

\paragraph{Group Size.}

After optimizing the cross-layer compression method, another factor is the group size.
% To investigate the effects of layer grouping, we partition the 32 layers into groups based on different grouping strategies. The parameter $k$ indicates that we store and share the quantized cache only in the $k$-th layer of each group.
To investigate the effects of layer grouping, we partition the total $L$ layers of a model (where $L=32$ for Mistral-7B and Llama 2-7B) into $L/G$ contiguous groups of size $G$. The parameter $k$ indicates that we store and share the quantized cache only in the $k$-th layer of each group.
We evaluate group sizes $G \in \{2,3,4\}$. This range is motivated by the empirical observation that while adjacent layers exhibit high similarity in their quantized representations (i.e., $G=2$, as shown in Figure~\ref{fig:delta}), this similarity diminishes gradually for layer distances greater than three.
For models with $L=32$ layers, $G=4$ thus serves as a sufficient upper bound for investigation due to this diminishing similarity.
We set all configurations under the same compression ratio, namely keep all layers in key cache and 20 layers in value cache based on KIVI-2bit framework, using Mistral-7b-instruct-v0.2.
As shown in \tref{group}, the model achieves the best performance with the configuration of \(G = 2\) and \(k = 0\).

% \vspace{10pt}
\begin{table}
% \small
  \centering
  \scalebox{0.92}{
    \begin{tabular}{llll}
    \hline
    \textbf{Method} & \textbf{Bit-width} & \textbf{TREC} & \textbf{SAMS} \\  \hline
    Full Cache      & 16                 & 71            & 42.75         \\
    KIVI            & 2                  & 71            & 43.7          \\
    AsymKV          & 1.5                & 70.5          & 43.44         \\  \hline
    AsymKV          & 1.375              & 69.5          & 42.76         \\
    XQuant          & 1.375              & \textbf{71.5}          & \textbf{45.18}         \\  \hline
    AsymKV          & 1.28               & 58.5          & 37.41         \\
    XQuant          & 1.28               & \textbf{68.5}          & \textbf{39.84}         \\  \hline
    AsymKV          & 1.15625            & 41            & 23.47         \\
    XQuant          & 1.15625            & \textbf{68.5 }         & \textbf{39.47    }
    \\
    \hline
    
    \end{tabular}
    }
  \caption{\label{tab:limit}
    The comparison of different configurations under extremely-low compression ratio.
  }
  \vspace{-10pt}
\end{table}

\paragraph{Performance-Compression Trade-offs.}

% To push the low-bit quantization to the limit, we conduct our XQuant in TREC and SAMSum subsets from LongBench, with different configurations under extremely-low compression ratio using Mistral-7b-instruct-v0.2.
% As shown in \tref{limit}, our XQuant can still maintaining competitive accuracy in these tasks even at an extreme 1.15625-bit setting, where it still retains over 90\% of the full floating-point baseline performance.
% To explore the limits of low-bit quantization, we evaluate XQuant on the TREC and SAMSum subsets from LongBench using Mistral-7b-instruct-v0.2 under extremely low compression ratios. As shown in Table~\ref{tab:limit}, XQuant maintains competitive accuracy even at an ultra-low 1.15625-bit setting, achieving over 90\% of the full-precision floating-point baseline performance.

% Table \ref{tab:limit}

Table \ref{tab:limit} evaluates the trade-offs between bit-width reduction and performance degradation across different quantization methods. As shown in Table \ref{tab:limit}, XQuant consistently outperforms other methods at the same bit-width, achieving higher scores on both TREC and SAMS benchmarks. Notably, even at an extremely low bit-width of 1.15625, XQuant preserves a significant portion of the model's performance, maintaining a TREC score of 68.5 compared to the full-cache baseline of 71.
These results demonstrate that XQuant effectively balances performance retention and compression, achieving state-of-the-art trade-offs in ultra-low-bit KV cache quantization.

% \paragraph{Efficiency Comparison.}

% To evaluate the reduction of GPU memory and improvement of throughput, we fix the input and output length to 161 and 338, and increase the batch size until exhausting the memory of a single NVIDIA GeForce RTX 3090 GPU (24GB).

% As shown in \fref{efficiency}, 

% // TODO: memory and throughput

%% file: pages/5-conclusion.tex
\section{Conclusion}

To alleviate the growing memory overhead in LLM inference, we propose XQuant, a plug-and-play framework that quantizes KV cache at an extreme compression ratio. Based on our observations on classical training-free quantization and the distributions of quantized integers, we propose a data-free calibration method and a compute-efficient cross-layer compression method.
Extensive experiments show that XQuant achieves state-of-the-art trade-offs between performance degradation and compression ratio, without sacrificing computational efficiency. Integrating these two novel methods, our XQuant achieves comparable performance with full-precision baseline under 1.4-bit quantization, and still maintains competitive performance for some tasks around an extremely 1.16-bit quantization.
% By offering up to (4×) larger batch sizes and (4×) throughput, XQuant paves the way for more cost-effective and scalable LLM deployments.

%% file: pages/6-tmp.tex
% \section*{Limitation}
% While XQuant demonstrates promising results across a range of representative models and benchmarks, future work may further validate its robustness and generalizability by extending evaluations to larger-scale models and more diverse downstream scenarios. Such expansion, given sufficient time and computational resources, would help reinforce the applicability of XQuant in broader real-world deployments.

\section*{Limitations and Future Work}

Our work presents several avenues for future exploration.
First, while XQuant demonstrates promising results on representative models and benchmarks, its robustness and generalizability could be further validated by extending evaluations to a wider range of newer-generation or larger-scale models and more diverse downstream scenarios.
Second, our current work relies on task-specific configurations. Although a unified setting proves robust (as shown in Appendix~\ref{sec:hyperparameter}), the development of an automated method to search for optimal configurations presents a valuable direction for future research.
Finally, the key innovations of XQuant — Data-Free Calibration and Cross-layer Compression — are in principle orthogonal to other KV cache compression paradigms. A fruitful area for future work would be to investigate their compatibility and potential synergies with these existing methods, potentially yielding even greater efficiency gains.

\section*{Acknowledgements}

This work was supported by the National Natural Science Foundation of China (Grant No. 62306216) and the Natural Science Foundation of Hubei Province of China (Grant No. 2023AFB816).

Hai Zhao's contribution was funded by the Major Program of the Chinese National Foundation of Social Sciences under Grant "The Challenge and Governance of Smart Media on News Authenticity" [No. 23\&ZD213].

The authors also gratefully acknowledge support from the Xiaomi Open-Competition Research Program.

%% file: pages/7-appendix.tex
\section{Preliminary Study on Relaxed-Contraint Mapping}

\begin{table}
  \centering
  \small
  \setlength{\extrarowheight}{2pt}
  \begin{tabular}{llccc}
    \hline
    \textbf{Method} &
    \textbf{Bit-width} &
    \textbf{\(\eta_1\)} &
    \textbf{\(\eta_2\)} &
    \textbf{MFQA-Zh}
    \\
    \hline
    Full Cache & 16 & / & / & 48.26  \\
    \hline
    KIVI & 2 & / & 0 & 42.27 \\
    KIVI & 2  & / & 0.05 & \textbf{44.34}  \\
    \hline
    AsymKV & 1.5 & 0 & 0 & 36.30\\
    AsymKV & 1.5 & 0 & 0.05 & \textbf{41.28} \\
    AsymKV & 1.5 & 0.2 & 0 & \textbf{42.78}   \\
    AsymKV & 1.5 & 0.2 & 0.05 & \textbf{43.81} \\
    \hline
  \end{tabular}
  \caption{\label{tab:pre}
    The comparison using different quantization methods with and without our calibration method in MultiFieldQA-Zh tasks from LongBench.
    % \e{i} = 0 means the i-bit quantization is deployed without calibration.
  }
  
\end{table}

As demonstrated in Figure \ref{fig:test}, the traditional quantization workflow faces higher quantization error in low-bit scenarios. In Section \ref{sec:calibration}, we propose a flexible mapping to mitigate the quantization error in this aspect.
Moreover, to provide empirical evidence supporting the effectiveness of the flexible mapping in the proposed calibration method, we employ its generalized form and conduct a preliminary study on the default KIVI-2bit and AsymKV-32/0 configurations. We extend this approach to a generalized B-bit quantization mechanism, where \(\eta_B\) serves as the corresponding parameter. Notably, when \(\eta_B =0\), the B-bit quantization operates without the flexible mapping.

The results in Table \ref{tab:pre} demonstrate that incorporating the flexible mapping function enhances model performance across different quantization settings.

\section{Preliminary Experiment on Layer-Wise Asymmetric Quantization}

In the existing method \cite{tao2024asymkv}, the KV cache for each layer is quantized using either 1-bit or 2-bit precision. A straightforward strategy to maximize the compression ratio is to apply 1-bit quantization to a greater number of layers.

However, a significant bottleneck arises, as it is nearly impossible to quantize the key cache at 1-bit precision without compromising performance. As shown in Table~\ref{tab:obs}, further compression by increasing the number of 1-bit quantized key cache layers is not feasible, as it leads to substantial performance degradation. This observation motivates us to explore alternative compression methodologies.

\begin{table}
    \small
  \centering
  \setlength{\extrarowheight}{1pt}
  \scalebox{0.85}{
  
  \begin{tabular}{llcc}
\hline
\textbf{Method} & \textbf{Bit-width} & \textbf{\# Key Layers in 1-bit} & \textbf{MFQA-Zh} \\ \hline
Full Cache      & 16                 & /                            & 48.26            \\
KIVI (32/32)    & 2                  & 0                            & 42.27            \\ \hline
AsymKV-24/32    & 1.875              & 8                            & 37.10            \\
AsymKV-16/32    & 1.75               & 16                           & 21.36            \\
AsymKV-8/32     & 1.625              & 24                           & 13.16            \\
AsymKV-0/32     & 1.5                & 32                           & 7.66  \\     \hline
\end{tabular}
}
  \caption{\label{tab:obs}
    Evaluation on LongBench based on AsymKV shows that the key cache is nearly impossible to quantized under 1-bit.
  }
\end{table}

\section{Equivalent Bit-width Analysis}
\label{app:parameter}

% In this section, we analyze the compression ratio in different compression methods.

Formally, let \(b, h, s, d\) be the batch size, the number of heads in GQA \cite{ainslie-etal-2023-gqa}, the sequence length and the dimension per head. The original \(L\) layers of KV cache occupies \(2L*bhsd * 16 \ bit\), which equals to \(2L*n*16 \ bit\) if we set \(n = bhsd\) for convenience.
 
Consider a typical KV cache quantization scheme \cite{Liu2024kivi}.
If we quantize all \(L\) layers of key cache and value cache into \(b\)-bit, the quantized KV cache memory usage is \(2L*n * b \ bit\).
\citealp{tao2024asymkv} uses a asymmetrical configurations for key and value caches across different layers. In their paper, Asym-\(l_k\)/\(l_v\) means quantizing the initial \(l_k\) layers of key cache and \(l_v\) of value cache into 2-bit, and quantizating 1-bit for others. So the quantized KV cache memory usage is \((2*l_k+(32-l_k)+2*l_v+(32-l_v)) * n \ bit\).
For example, Asym-1.5bit stands for Asym-32/0 in our paper, which can be calculated to \(3L * n \ bit\) and can be equivalently considered as a 1.5-bit symmetrical quantization for better understanding of the compression ratio.
% The related parameters in XQuant are \(kq, vq, km, vm\). The equivalent bit-width \(B\)  can be expressed as follows:
% \(B=((32-max(kq,km))/2+(max(kq,km)-min(kq,km))+(max(kq,km)+min(kq,km))*2 + (32-max(vq,vm))/2+(max(vq,vm)+min(vq,vm))+(max(vq,vm)+min(vq,vm))*2)/64\).
% In a classical configuration in our paper \(kq=30,vq=2,km=32,vm=16\), in Key cache we apply:
% \begin{itemize}
%     \item Quantization: 2-bit for layers [0,kq) and 1-bit for layers [kq,32).
%     \item Cross-layer compression: Applied to layers [km,32).
% \end{itemize}
% The same applies to Value cache. The equivalent bit-widths are computed as follows:
% \newline
% \newline
% \(Key: \frac{(32-30)+30*2}{32} = 1.9375\)
% \newline
% \newline
% \(Value: \frac{(32-16)/2+(16-2)+2*2}{32} = 0.8125\)

% The average bit-width is therefore 1.375, which appears as 1.38 in most parts of this paper. We have already provided the full parameter sets used in our experiments are listed in Appendix \ref{sec:config}.

The related parameters in XQuant are \(kq\), \(vq\), \(km\), and \(vm\). The equivalent bit-width \(B\) can be expressed as follows:
% \[
% B = \frac{1}{64} \left( \left( 32 - \max(kq, km) \right) / 2 + \left( \max(kq, km) - \min(kq, km) \right) + \left( \max(kq, km) + \min(kq, km) \right) \times 2 + \left( 32 - \max(vq, vm) \right) / 2 + \left( \max(vq, vm) + \min(vq, vm) \right) + \left( \max(vq, vm) + \min(vq, vm) \right) \times 2 \right)
% \]
\(B=((32-max(kq,km))/2+(max(kq,km)-min(kq,km))+(max(kq,km)+min(kq,km))*2 + (32-max(vq,vm))/2+(max(vq,vm)+min(vq,vm))+(max(vq,vm)+min(vq,vm))*2)/64\).

In the classical configuration in our paper, \(kq = 30\), \(vq = 2\), \(km = 32\), and \(vm = 16\), in key cache we apply 2-bit quantization to the layers \([0, kq)\) and 1-bit quantization to the layers \([kq, 32)\), and cross-layer compression to the layers \([km, 32)\). The value cache is processed in the same manner. Therefore, the equivalent bit-widths of the key and value caches are computed as follows:

\[
B_{k} = \frac{(32 - 30) + 30 * 2}{32} = 1.9375
\]

\[
B_{v} = \frac{(32 - 16) / 2 + (16 - 2) + 2 * 2}{32} = 0.8125
\]

The average bit-width is therefore 1.375, which appears as 1.38 in most parts of this paper. More parameter sets used in our experiments are listed in Appendix \ref{sec:config}.

% For clarity, we note that for all per-group quantization methods discussed, the reported "equivalent bit-width" exclusively accounts for the memory footprint of the quantized integer tensors. The memory overhead associated with storing metadata, such as scaling factors and zero-points, is not included in this calculation. This approach is intentionally adopted to maintain methodological consistency with seminal works in the field (e.g., KIVI \cite{Liu2024kivi} and AWQ \cite{lin2024awq}), thereby facilitating a direct and fair comparison. Furthermore, we maintain that the comparisons within this paper are rigorous, as all quantization methods were implemented with identical group sizes and residual lengths. Consequently, the unaccounted metadata overhead is uniform across all tested methods and does not affect their relative performance rankings.

To maintain consistency with seminal works (e.g., KIVI \cite{Liu2024kivi} and GPTQ \cite{frantar2022gptq}), our reported "equivalent bit-width" for asymmetrical quantization methods considers only the quantized integer tensors, excluding metadata overhead like scaling factors and zero-points. The comparisons remain rigorous, as all evaluated quantization methods were implemented with identical group sizes and residual lengths. This ensures the unaccounted overhead is uniform across all methods and does not affect their relative performance rankings.

\begin{figure}
    \centering
    \includegraphics[width=0.75\linewidth]{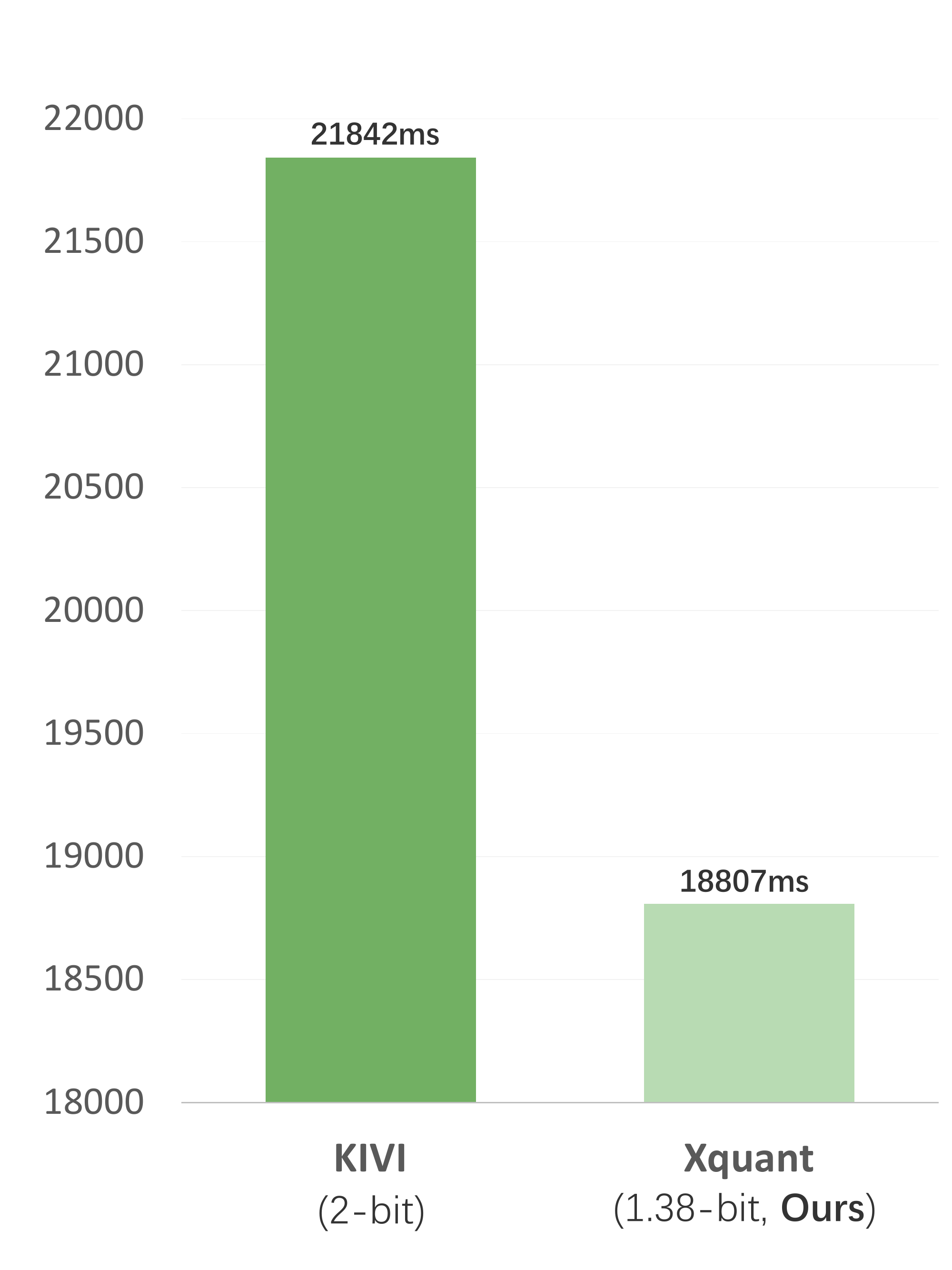}
    \caption{Comparison of Execution Time.}
    \label{fig:time}
\end{figure}

\section{Efficiency analysis}

Using Mistral-7B as an example, we theoretically analyze the computational cost of our two key improvements. During the \textbf{calibration} step, generating each token incurs only 64 additional floating-point multiplications and 32 additions (Equation \ref{eq:12}), which are negligible in practice. Moreover, as described in Section \ref{sec:speedup}, the \textbf{cross-layer compression} step optimizes efficiency by skipping certain parts of the quantization process (Equation \ref{eq:2}).

To evaluate inference efficiency, we adopt the same experimental setup as implemented in KIVI's repository, using a batch size of 16, a prompt length of 1024, and an output length of 128. As shown in Figure \ref{fig:time}, XQuant, by leveraging its unique speedup mechanism, demonstrates competitive inference efficiency.

\section{Hyperparameter}
\label{sec:hyperparameter}

The related parameters in XQuant are \(kq\), \(vq\), \(km\), and \(vm\).
In XQuant, we quantize the lower \(kq\), \(vq\) layers of key and value cache into 2-bit, while quantizing others into 1-bit. We apply cross-layer compression from the \(km\) th, \(vm\) th layer of key and value cache. All the configurations are summarized in \tref{param}.

As demonstrated in Table~\ref{tab:hyperparameters}, additional experiments on the Mistral-7B-Instruct model using the LongBench benchmark show that XQuant, with a fixed $\eta_1 = 1/6$ and $\eta_2 = 0.045$, consistently delivers strong performance as well. These results suggest that this fixed set of hyperparameters are robust and can generalize effectively across different datasets. Therefore, task-specific hyperparameter tuning is superior but not necessary, and the method can achieve reliable performance with a fixed, pre-selected set of hyperparameters.

% We will further refine the explanation of this point in the revised version of our manuscript to provide clearer insights into the practicality and robustness of the approach.

\begin{table*}
\centering
\resizebox{\textwidth}{!}{
\begin{tabular}{lllcc ccccc c}
\hline
\textbf{Method} & \textbf{Bit-width} & \textbf{Hyperparameters} & \textbf{HQA} & \textbf{2Wiki} & \textbf{MSQ} & \textbf{TREC} & \textbf{TQA} & \textbf{SAMS} & \textbf{PC} & \textbf{Avg} \\ \hline
Full Cache       & 16                 & /                       & 43.02       & 27.10         & 18.78        & 71.00         & 86.23         & 42.75         & 2.75         & 41.66        \\
AsymKV           & 1.5                & /                       & 37.17       & 22.77         & 15.76        & 70.50         & 86.25         & 43.44         & 3.16         & 39.86        \\ \hline
XQuant           & 1.38               & Task-specific           & 42.90       & 26.65         & 17.44        & 71.50         & 84.50         & 45.18         & 5.71         & 41.98        \\
XQuant           & 1.38               & \textbf{Static}         & \textbf{42.64}  & \textbf{25.16} & \textbf{16.91} & \textbf{70.50}  & \textbf{84.50} & \textbf{42.64} & \textbf{4.57}  & \textbf{40.99}  \\ \hline
\end{tabular}
}
\caption{Evaluation of different KV cache compression methods using static hyperparameters setting.}
\label{tab:hyperparameters}
\end{table*}

\section{Cross-Layer Compression Strategy}
\label{app:int}

Under 2-bit quantization, the values in the KV cache are restricted to the discrete integer set \(\{i \in \mathbf{Z} \mid 0 \le i \le 3\}\). Therefore, a rounding operation is required after weighted averaging. If standard rounding-to-nearest is applied, the range of \(\gamma_0\) can be divided into six disjoint intervals, as summarized in Table~\ref{tab:gamma}. The derivation is as follows:

Let \(e_0\) and \(e_1\) denote the \(B\)-bit quantized values at the same position in adjacent layers of \(\mathbf{X_Q}\). Then the merged value \(e_m\) after cross-layer compression is computed as:
\begin{align*}
    e_m 
    &= \left\lfloor \frac{\gamma_0 e_0 + \gamma_1 e_1}{\gamma_0 + \gamma_1} \right\rceil \\
    &= \left\lfloor \gamma_0 e_0 + (1-\gamma_0) e_1 \right\rceil \\
    &= e_1 + \left\lfloor \gamma_0 (e_0 - e_1) \right\rceil.
\end{align*}

Without loss of generality, assume \(e_0 \ge e_1\) and define \(\delta = e_0 - e_1 \ge 0\). Then we have:
\begin{equation}
\label{eq:54}
    e_m = e_1 + \left\lfloor \gamma_0 \delta \right\rceil,
\end{equation}
where \(\gamma_0 \in [0,1]\) and \(\delta \in \mathbf{Z} \cap [0,3]\). Since \(\gamma_0 \delta \in [0,\delta]\), the rounding term \(\left\lfloor \gamma_0 \delta \right\rceil\) in Eq.~\ref{eq:54} can only take \(\delta + 1\) discrete values.
Let \(\left\lfloor \gamma_0 \delta \right\rceil = c\), where \(c \in \mathbf{Z} \cap [0,\delta]\). Then:
\begin{equation}
    \gamma_0 \delta \in \left(c - \frac{1}{2}, c + \frac{1}{2} \right) \cap [0, \delta],
\end{equation}
which yields the following constraint for \(\gamma_0\), when \(\delta > 0\):
\begin{equation}
\label{eq:56}
    \gamma_0 \in \left( \frac{c - 1/2}{\delta}, \frac{c + 1/2}{\delta} \right) \cap [0,1].
\end{equation}

We now enumerate all valid combinations of \(\delta\) and \(c\) from Equation \ref{eq:56}:

\begin{itemize}
    \item \(\delta = 0\): Only one possible value exists; trivial case omitted.
    \item \(\delta = 1\): 
    \begin{itemize}
    \renewcommand\labelitemii{\textbullet}
        \item \(c = 0\): \(\gamma_0 \in [0, 1/2)\)
        \item \(c = 1\): \(\gamma_0 \in (1/2, 1]\)
    \end{itemize}
    \item \(\delta = 2\): 
    \begin{itemize}
    \renewcommand\labelitemii{\textbullet}
    
        \item \(c = 0\): \(\gamma_0 \in [0, 1/4)\)
        \item \(c = 1\): \(\gamma_0 \in (1/4, 3/4)\)
        \item \(c = 2\): \(\gamma_0 \in (3/4, 1]\)
    \end{itemize}
    \item \(\delta = 3\): 
    \begin{itemize}
    \renewcommand\labelitemii{\textbullet}
    
        \item \(c = 0\): \(\gamma_0 \in [0, 1/6)\)
        \item \(c = 1\): \(\gamma_0 \in (1/6, 1/2)\)
        \item \(c = 2\): \(\gamma_0 \in (1/2, 5/6)\)
        \item \(c = 3\): \(\gamma_0 \in (5/6, 1]\)
    \end{itemize}
\end{itemize}

Collectively, this yields six effective intervals of \(\gamma_0\), as summarized in Table~\ref{tab:gamma}.

% It is worth noting that when \(\gamma_0 \in [0, 1/6)\) or \(\gamma_0 \in (5/6, 1]\), the compression result becomes equivalent to the accelerated strategy defined by Eq.~\ref{eq:47}, where the selector \(k\) takes value 0 or 1. This simplification can substantially reduce redundant computation and accelerate inference.

% Furthermore, for edge cases where \(\gamma_0 \in \{1/6, 1/4, 1/2, 3/4, 5/6\}\), standard rounding may lead to ambiguity. Among these, \(\gamma_0 = 1/2\) is a particularly important scenario for examining the impact of rounding strategies such as floor vs. ceil. Comparative results for different cross-layer compression strategies are reported in Table~\ref{tab:gamma}.

\section{Comparison with Other Cross-Layer Compression Methods}
\label{sec:comparison}

Several prior works have explored inter-layer redundancy from different perspectives. To eliminate potential confusion, we clarify several key distinctions and highlight innovations as follows:
(\textbf{a}) Most existing methods compute KV caches at a subset of layers. However, these approaches require additional training steps and, in some cases, even full retraining, significantly limiting scalability. In contrast, XQuant is designed as a plug-and-play solution that leverages deeper insights to enable effective redundancy reduction without any additional training.
(\textbf{b}) XQuant is the only method that explicitly considers inter-layer redundancy through the lens of quantization. After quantization, the KV cache is decomposed into three components: the quantized cache, zero-points, and scaling factors. We demonstrate that the quantized cache, consisting solely of integers, exhibits substantial inter-layer similarity. Meanwhile, the zero-points and scaling factors, which require minimal storage, are retained individually to preserve per-layer characteristics without being compressed.
(\textbf{c}) MiniCache \cite{liu2024minicache} is another training-free method that primarily introduces a retention-recovery mechanism for cache magnitudes and unmergable tokens. However, such operations are not directly compatible in mainstream open-source KV quantization frameworks. Furthermore, its use of the SLERP function imposes several constraints, making it inapplicable to quantized caches, which fundamentally differs from XQuant.

\begin{table}
\centering
\resizebox{0.42\textwidth}{!}{
\begin{tabular}{llcc}
\hline
\textbf{Method}     & \textbf{Bit-width} & \textbf{2Wiki}  & \textbf{HQA}  \\ \hline
Full Cache & 16        & 58.20     & 61.88     \\
AsymKV     & 1.4       & 38.55     & 44.69     \\ \hline
XQuant     & 1.4       & \textbf{54.16} & \textbf{57.44} \\ \hline
\end{tabular}
}
\caption{
\label{tab:qwen}
Comparison of XQuant with Full Cache and AsymKV on the Qwen2.5-14B model using the LongBench benchmark.
}
\end{table}

\section{Evaluation on Qwen2.5-14B}

As shown in Table \ref{tab:qwen}, we evaluated XQuant on a larger-scale and newer-generation model, Qwen2.5-14B \cite{qwen2025qwen25technicalreport}, using the LongBench benchmark. The results demonstrate that XQuant generalizes well to different models, maintaining a superior trade-off between model performance and compression ratio.

\begin{table*}
  \centering
  % \small
  % \setlength{\extrarowheight}{3pt}
    \begin{tabular}{llllllll}
    \hline
    Model                                     & Dataset    & kq & vq & km & vm & eta1 & eta2  \\
    \hline
    \multirow{1}{*}{Mistral-7b-v0.3}         & TruthfulQA & 30 & 2  & 32 & 16 & 0    & 0     \\
        \hline
    \multirow{7}{*}{Mistral-7b-instruct-v0.2} & HQA        & 30 & 2  & 32 & 16 & 1/6  & 0.045 \\
                                              & 2Wiki      & 32 & 0  & 32 & 16 & 0  & 0.09  \\
                                              & MSQ        & 32 & 0  & 32 & 16 & 1/6  & 0     \\
                                              & TREC       & 30 & 2  & 32 & 16 & 1/6  & 0     \\
                                              & TQA        & 30 & 2  & 32 & 16 & 1/6  & 0.09  \\
                                              & SAMS       & 30 & 2  & 32 & 16 & 0    & 0     \\
                                              & PC         & 32 & 0  & 32 & 16 & 0    & 0.045 \\
    \hline
    \multirow{1}{*}{Llama2-7b}               
                                              & TruthfulQA & 28 & 0  & 32 & 28 & 1/3  & 0     \\
    \hline
    \multirow{7}{*}{Llama2-7b-chat}           & HQA        & 28 & 0  & 32 & 28 & 1/6  & 0.045     \\
                                              & 2Wiki      & 28 & 0  & 32 & 28 & 1/3  & 0.045     \\
                                              & MSQ        & 28 & 0  & 32 & 28 & 1/3  & 0     \\
                                              & TREC       & 32 & 0  & 32 & 20 & 1/6  & 0     \\
                                              & TQA        & 32 & 0  & 32 & 20 & 1/6  & 0     \\
                                              & SAMS       & 32 & 0  & 32 & 20 & 0    & 0     \\
                                              & PC         & 32 & 0  & 32 & 20 & 1/3  & 0.045 \\
    \hline
    \end{tabular}
  \caption{\label{tab:param}
    The configurations of our main experiments.
  }
\end{table*}

\section{Configurations}
\label{sec:config}

The Configurations of XQuant in our main experiments are summarized in \tref{param}

\section{XQuant Pseudo Code} \label{sec:algorithms}
The pseudo code for the whole workflow is provided in Algorithm \ref{alg:workflow1} and \ref{alg:workflow2}.

% \begin{algorithm}
% \caption{XQuant Procedure}
% \KwIn{$kq, vq, km, vm, \eta[2]$}
% \KwOut{Optimized Quantized Cache}

% \For{$l \gets 0$ \textbf{to} $31$}{
%     \eIf{$l < vm$ \textbf{or} $l \bmod 2 = 0$}{
%         $\text{KeyCache}[l] \gets \textbf{Quantize}(X^l_k, 2 \textbf{ if } l < kq \textbf{ else } 1)$\;
%     }{
%         $\text{KeyCache}[l] \gets \textbf{PseudoQuantize}(X^l_k, 2 \textbf{ if } l < kq \textbf{ else } 1)$\;
%     }
%     \eIf{$l < vm$ \textbf{or} $l \bmod 2 = 0$}{
%         $\text{ValueCache}[l] \gets \textbf{Quantize}(X^l_v, 2 \textbf{ if } l < vq \textbf{ else } 1)$\;
%     }{
%         $\text{ValueCache}[l] \gets \textbf{PseudoQuantize}(X^l_v, 2 \textbf{ if } l < vq \textbf{ else } 1)$\;
%     }
% }

% \For{$l \gets 0$ \textbf{to} $31$}{
%     \eIf{$l < km$ \textbf{or} $l \bmod 2 = 0$}{
%         $\text{DequantizedKey} \gets \textbf{Dequantize}(
%         \text{KeyCache}[l][0], \text{KeyCache}[l][1], \text{KeyCache}[l][2])$\;
%     }{
%         $\text{DequantizedKey} \gets \textbf{Dequantize}(
%         \text{KeyCache}[l-1][0], \text{KeyCache}[l-1][1], \text{KeyCache}[l][2])$\;
%     }
%     \eIf{$l < vm$ \textbf{or} $l \bmod 2 = 0$}{
%         $\text{DequantizedValue} \gets \textbf{Dequantize}(
%         \text{ValueCache}[l][0], \text{ValueCache}[l][1], \text{ValueCache}[l][2])$\;
%     }{
%         $\text{DequantizedValue} \gets \textbf{Dequantize}(
%         \text{ValueCache}[l-1][0], \text{ValueCache}[l-1][1], \text{ValueCache}[l][2])$\;
%     }
% }
% \end{algorithm}

\clearpage

\begin{algorithm}

\caption{\label{alg:workflow1}XQuant Procedure}
\SetAlgoNoEnd % 不显示 'end' 关键字
\DontPrintSemicolon % 不在每行末尾打印分号
\SetInd{0.5em}{0.5em} % 设置缩进大小
\SetNlSty{}{}{} % 行号样式，无特殊格式
\SetKwInOut{KwIn}{Input}
\SetKwInOut{KwOut}{Output}

% 调整算法环境的左右边距

\KwIn{$kq$, $vq$, $km$, $vm$, $\eta[2]$}
\KwOut{Optimized Quantized Cache}

\For{$l \gets 0$ \textbf{to} $31$}{
    \eIf{$l < vm$ \textbf{or} $l \bmod 2 == 0$}{
        $\text{KeyCache}[l] \gets \textbf{Quantize}\big(X^l_k,\ 2\ \textbf{if } l < kq\ \textbf{else } 1\big)$\;
    }{
        $\text{KeyCache}[l] \gets \textbf{PseudoQuantize}\big(X^l_k,\ 2\ \textbf{if } l < kq\ \textbf{else } 1\big)$\;
    }
    \eIf{$l < vq$ \textbf{or} $l \bmod 2 == 0$}{
        $\text{ValueCache}[l] \gets \textbf{Quantize}\big(X^l_v,\ 2\ \textbf{if } l < vq\ \textbf{else } 1\big)$\;
    }{
        $\text{ValueCache}[l] \gets \textbf{PseudoQuantize}\big(X^l_v,\ 2\ \textbf{if } l < vq\ \textbf{else } 1\big)$\;
    }
}

\For{$l \gets 0$ \textbf{to} $31$}{
    \eIf{$l < km$ \textbf{or} $l \bmod 2 == 0$}{
        $\text{DequantizedKey} \gets \textbf{Dequantize}\big($ \\
        \Indp % 增加缩进
        $\text{KeyCache}[l][0],$ \\
        $\text{KeyCache}[l][1],$ \\
        $\text{KeyCache}[l][2]\big)$\;
        \Indm % 减少缩进
    }{
        $\text{DequantizedKey} \gets \textbf{Dequantize}\big($ \\
        \Indp
        $\text{KeyCache}[l-1][0],$ \\
        $\text{KeyCache}[l-1][1],$ \\
        $\text{KeyCache}[l][2]\big)$\;
        \Indm
    }
    \eIf{$l < vm$ \textbf{or} $l \bmod 2 == 0$}{
        $\text{DequantizedValue} \gets \textbf{Dequantize}\big($ \\
        \Indp
        $\text{ValueCache}[l][0],$ \\
        $\text{ValueCache}[l][1],$ \\
        $\text{ValueCache}[l][2]\big)$\;
        \Indm
    }{
        $\text{DequantizedValue} \gets \textbf{Dequantize}\big($ \\
        \Indp
        $\text{ValueCache}[l-1][0],$ \\
        $\text{ValueCache}[l-1][1],$ \\
        $\text{ValueCache}[l][2]\big)$\;
        \Indm
    }
}
\end{algorithm}

\begin{algorithm}
\caption{\label{alg:workflow2}Supporting Functions}
\SetKwFunction{PseudoQuantize}{PseudoQuantize}
\SetKwFunction{Quantize}{Quantize}
\SetKwFunction{Dequantize}{Dequantize}
\SetKwFunction{Calibrate}{Calibrate}

\SetKwProg{Fn}{Function}{:}{}

\Fn{\PseudoQuantize{$X$, $n\_bits$}}{
    $zero\_point \gets \min(X)$ \tcp*[h]{Find the minimum value of $X$}\;
    $scaling\_factor \gets \frac{\max(X) - \min(X)}{2^{n\_bits} - 1}$ 
    \tcp*[h]{Calculate scaling factor}\;
    \Return \\
    \Indp % 增加缩进
    $\textbf{Calibrate}(zero\_point, \, $
    $scaling\_factor, \, n\_bits)$, \\
    None\;
    \Indm % 恢复缩进
}

\Fn{\Quantize{$X$, $n\_bits$}}{
    $zero\_point \gets \min(X)$\;
    $scaling\_factor \gets \frac{\max(X) - \min(X)}{2^{n\_bits} - 1}$\;
    $quantized\_cache \gets 
    \textbf{round}\left(\frac{X - zero\_point}{scaling\_factor}\right)$ 
    \tcp*[h]{Round to nearest quantized value}\;
    \Return \\
    \Indp
    $\textbf{Calibrate}(zero\_point, \, $
    $scaling\_factor, \, n\_bits)$, \\
    $quantized\_cache$\;
    \Indm
}

\Fn{\Dequantize{$zero\_point$, $scaling\_factor$, $quantized\_cache$}}{
    \Return $quantized\_cache \cdot scaling\_factor + zero\_point$
    \tcp*[h]{Reconstruct original value}\;
}

\Fn{\Calibrate{$zero\_point$, $scaling\_factor$, $n\_bits$}}{
    $zero\_point\_cali \gets zero\_point + 
    scaling\_factor \cdot \eta[n\_bits]$ 
    \tcp*[h]{Adjust zero point based on $\eta$}\;
    
    $scaling\_factor\_cali \gets scaling\_factor \cdot 
    \big(1 - 2 \cdot \eta[n\_bits]\big)$ 
    \tcp*[h]{Adjust scaling factor based on $\eta$}\;
    
    \Return \\
    \Indp
    $zero\_point\_cali, \, scaling\_factor\_cali$
    \tcp*[h]{Return calibrated values}\;
    \Indm
}

\end{algorithm}